\documentclass[letterpaper, 10 pt, conference]{ieeeconf}  % Comment this line out if you need a4paper

\IEEEoverridecommandlockouts                              % This command is only needed if 
\usepackage{graphicx}
\usepackage{amsmath}
\usepackage{amssymb}
\usepackage{wrapfig}
\usepackage{placeins}
\usepackage{hyperref}
\usepackage{subfig}
\usepackage{tabu}
\usepackage{subfig}
\usepackage{booktabs}% for better rules in the table
\usepackage{float}

\usepackage{array}
\newcolumntype{P}[1]{>{\centering\arraybackslash}p{#1}}

\title{\LARGE \bf
An A* Curriculum Approach to Reinforcement Learning for RGBD Indoor Robot Navigation 
}

\author{Kaushik Balakrishnan$^{1}$, Punarjay Chakravarty$^{1}$ and Shubham Shrivastava$^{1}$% <-this % stops a space
\thanks{$^{1}$ The authors are with Ford Greenfield Labs, Palo Alto, CA, USA.
        {\tt\small kbalak18@ford.com},{\tt\small \, pchakra5@ford.com},{\tt\small \, sshriva5@ford.com}}%
}

\begin{document}
% Note: This paper has to be trimmed to 6 pages + references for the IROS submission.
% List of things to remove for the IROS paper:
% 1. Combine figures 2 & 3, maybe remove 1. JAY FIXED THIS 
% 2. Add a figure for the 1st page which shows habitat environment top-down view and sensor view JAY FIXED THIS 
% 3. Remove A* explanation - DONE, COMMENTED IT (KB)
% 4. Add Transfer Learning experiments
% 5. Add Hierarchical Reinforcement Learning reference - DONE (KB)
% 6. Possibly add PPO equations - I FEEL WE MUST HAVE THIS FOR ARXIV, NOT FOR IROS AS THERE IS A 6 PAGE LIMIT (KB)

\maketitle
\thispagestyle{empty}
\pagestyle{empty}

\begingroup
\setcounter{footnote}{1}%
\endgroup

%%%%%%%%%%%%%%%%%%%%%%%%%%%%%%%%%%%%%%%%%%%%%%%%%%%%%%%%%%%%%%%%%%%%%%%%%%%%%%%%
\begin{abstract}
Training robots to navigate diverse environments is a challenging problem as it involves the confluence of several different perception tasks such as mapping and localization, followed by optimal path-planning and control. Recently released photo-realistic simulators such as Habitat \cite{Habitat19} allow for the training of networks that output control actions directly from perception: agents use Deep Reinforcement Learning (DRL) to regress directly from the camera image to a control output in an end-to-end fashion. This is data-inefficient and can take several days to train on a GPU. Our paper tries to overcome this problem by separating the training of the perception and control neural nets and increasing the path complexity gradually using a curriculum approach. Specifically, a pre-trained twin Variational AutoEncoder (VAE) \cite{kingma2013auto} is used to compress RGBD (RGB \& depth) sensing from an environment into a latent embedding, which is then used to train a DRL-based control policy.
A*, a traditional path-planner is used as a guide for the policy and the distance between start and target locations is incrementally increased along the A* route, as training progresses. We demonstrate the efficacy of the proposed approach, both in terms of increased performance and decreased training times for the PointNav task in the Habitat simulation environment.
This strategy of improving the training of direct-perception based DRL navigation policies is expected to hasten the deployment of robots of particular interest to industry such as co-bots on the factory floor and last-mile delivery robots.\footnote{ More information and videos of robot navigation using our approach can be found at: \url{https://www.towardsautonomy.com/Robot-Navigation-Using-Vision-Embedding}}
\end{abstract}

\begin{figure}[h]
\begin{center}
    \includegraphics[width=0.9\linewidth]{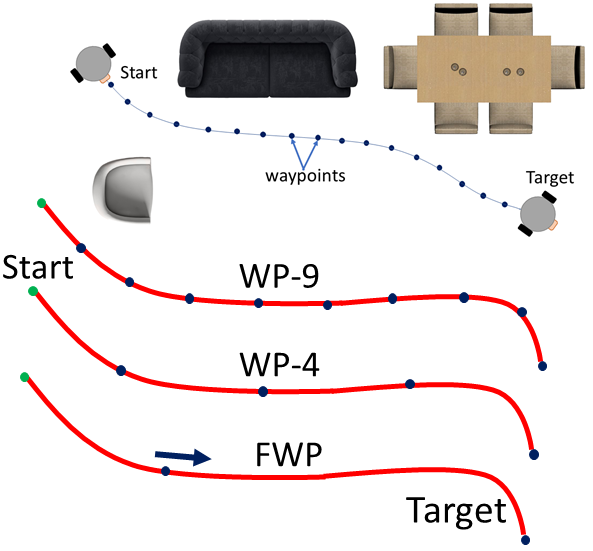}
\end{center}
% \vspace*{-0.55cm} 
   \caption{Top: waypoints generated between desired start and target locations by the A* algorithm. This work looks at assisting the training of a DRL-based robot navigation policy, by incrementally increasing the difficulty of the navigation task in a curriculum. Bottom: 3 curriculum training approaches with 9 and 4 discrete waypoints and a continuously moving waypoint: WP-9, WP-4 \& FWP.}
\label{fig:waypoints}
% \vspace*{-0.80cm} 
\end{figure}
%Jay TODO: Decrease all WP numbers by 1 in figure and caption.
% Increase size of font in inset.
% Remove numbering of waypoints in the robot diagram. Just call them waypoints.
% Change "goal" to "target"
% Try it as a double col figure instead of having an inset.
%Jay: Done.

%%%%%%%%%%%%%%%%%%%%%%%%%%%%%%%%%%%%%%%%%%%%%%%%%%%%%%%%%%%%%%%%%%%%%%%%%%%%%%%%
\section{INTRODUCTION}
\label{introduction}
To go from point A to B in an indoor environment is challenging for a mobile robot. In the absence of GPS and using the visual/RGBD sensor available on the robot, one has to map an environment \& localize in it (SLAM) and then path-plan an obstacle-free route to get from a start to target location. This was the traditional approach to mobile robotics. Recently, Deep Reinforcement Learning (DRL) has shown to provide more robust navigation policies compared to SLAM, if the robot (agent) is trained in simulation and exposed to an order of magnitude more experience \cite{Habitat19}.

% Learning robust navigation policies for indoor robot navigation is a challenging problem as it involves the confluence of several different tasks such as perception of the environment, localization, optimal path planning, collision avoidance, safety, etc. Each of these tasks poses difficulties, and errors in one sub-task can accrue and affect learning in others, and therefore breaking up these different tasks and learning them one at a time is a promising approach. 
This involves training navigation policies that regress directly from the camera image to a control output. However, splitting this task into two: learning a compact state representation, termed ``representation learning" and then using this representation to learn a robust control policy \cite{Grossberg1993, Lesort2018, Ha18, Splitnet19, Nachum19, Tschannen18, Nair18} has the following advantages:  (1) errors in policy learning will not affect perception as the latter is decoupled from the former, but the vice versa is not true; (2) once perception is learned, it can be reused to learn multiple policies for different tasks \cite{Splitnet19}, which is not feasible in complete end-to-end training as the perception needs to be re-learned every time a new task is learned. These advantages have the potential to speed-up the overall learning of the task at hand.

% The first task in a DRL pipeline is to learn a compact state representation, termed ``representation learning" and the second task is to use this representation to learn a robust control policy \cite{Grossberg1993, Lesort2018, Ha18, Splitnet19, Nachum19, Tschannen18, Nair18}. This has two advantages: (1) errors in policy learning will not affect perception as the latter is decoupled from the former, but the vice versa is not true; (2) once perception is learned, it can be reused to learn multiple policies for different tasks \cite{Splitnet19}, which is not feasible in complete end-to-end training as the perception needs to be re-learned every time a new task is learned. These advantages have the potential to speed-up the overall learning of the task at hand.      

The recently released Gibson \cite{Gibson} and Habitat simulators \cite{Habitat19} have generated excitement in the field of RGBD vision-based robot navigation in indoor environments. In SplitNet \cite{Splitnet19}, the authors investigated three tasks in the Habitat environment: (1) Point-to-Point Navigation (PointNav); (2) Scene Exploration; (3) Run Away from Location (Flee). They decoupled the perception and policy, and used an Encoder-Decoder architecture, where the visual/perception encoder is trained using auxiliary visual and motion-based tasks, and the policy Decoder is trained on embodied tasks using the visual Encoder. They demonstrated robust learning of both perception and policy on all three tasks, including transfer to new visual environments as well as to new embodied tasks. In \cite{WatkinsValls}, the authors undertook Imitation Learning to train a robot to navigate in the Gibson simulator using the Dijkstra algorithm and obtained high success rates. In another study, the authors used DRL for target-driven robot navigation in an indoor scenes simulator called AI2THOR \cite{AI2THOR}, where only RGB images of the state and the target are used to train the navigation policy network. All these studies demonstrated robustness of DRL for robot navigation using large amounts of vision data, however techniques to speed-up DRL for vision-based navigation is warranted as most of these techniques are GPU-compute intensive and their speed-up can help in faster learning and deployment of robots in the real world.    

% KB- THE ABOVE PARA HAS BEEN MODIFIED TO DISCARD THE VAE-BASED PAPERS AND REPLACING THEM WITH VISION-BASED DRL PAPERS

In this paper, we build on these past investigations and train DRL agents for the problem of indoor robot navigation in the Habitat environment by separating perception (i.e., representation learning) and control (i.e., navigation policy). We use a VAE to encode RGB and Depth images, and use these latent encodings as well as
%goal-based PointGoal sensor 
a reading and heading angle for the target (from the PointGoal sensor), to learn navigation policies. 

Additionally, we use a traditional path-planner, A* to assist the DRL agent during training. A* guides the agent by giving it shorter-distance goal locations (waypoints) between the original start and target locations.
We experiment with two different curriculum-based training of the DRL agents, one by decreasing the number of intermediate waypoints used (termed the SWP-N agent) or by moving the episodic goal farther away from the start position (termed the FWP agent). 
We describe the problem and our method in Section \ref{problem_setup}, implementation details in Section \ref{implementation_details} and an experimental analysis of these methods in Section \ref{experimental_results}.

In summary, our contributions are as follows: (1) a principled approach to compare different navigation-agnostic VAE-based perception embeddings for their usefulness to a DRL in learning a subsequent navigation policy; (2) Using a traditional  A* path-planning algorithm in a curriculum fashion to assist in the training process of this navigation policy.

% (2) DRL-based navigation policy with LSTM layers using the learned perception embedding as input; (3) using a traditional A* path-planning algorithm to assist in the training process of the navigation policy; and (4) training DRL agents using a curriculum. 

\section{RELATED WORK}
\label{related_work}
% This work investigates the separation of perception and control for navigation of an indoor robot and using a curriculum to learn the navigation control. We also use a traditional path-planning approach, A* with modern Deep Reinforcement Learning (DRL) algorithms for efficient navigation in an indoor setting.
%Jay: above para not required, as we have just explained this in the previous para at end of intro.
\begin{figure}[h]
\begin{center}
    \includegraphics[width=0.98\linewidth]{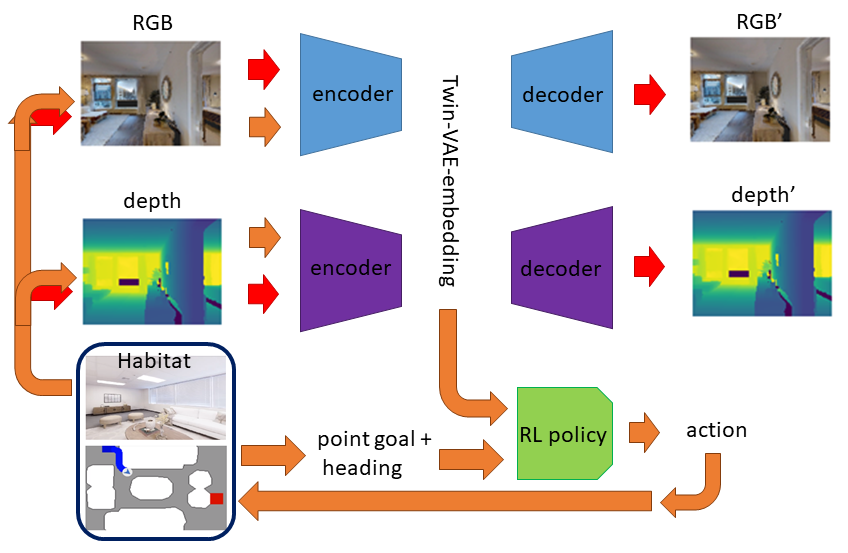}
\end{center}
% \vspace*{-0.55cm} 
   \caption{A twin (RGB-depth) VAE learns an embedded representation of the environment, which is then used to train a navigation policy using DRL. Information flow during the VAE and DRL training are shown in red and orange respectively.}
\label{fig:vaeRL}
% \vspace*{-0.60cm} 
\end{figure}

\subsection{Visual perception}

Training end-to-end Vision-based DRL navigation policies can be very time consuming as the CNNs used to learn vision-based features involve several matrix operations, and this can particularly require several million images/experiences, accompanied by several days of compute hours, for training the DRL policy \cite{Habitat19, AI2THOR}. End-to-end learning of DRL policies from RGB images has been very successful for Atari games \cite{Mnih2013}, but for larger images such as the Habitat environment, it is very challenging \cite{Habitat19, DDPPO}. For learning robust navigation policies using RGBD images, it is critical to obtain good perception representation that is feature rich and compact enough. 

We use the Variational Auto-Encoder (VAE) \cite{kingma2013auto} for perception representation; VAE is a generative version of the vanilla Auto-encoder, that constrains the latent space $z$ to a Gaussian distribution with zero mean and unit variance $p(z) = \mathcal{N}(0,I)$. The encoder is used to produce the latent space encoding and the decoder takes in $z$ to reconstruct the input image. The VAE is trained using a combination of the reconstruction loss (typically, L2) and the KL divergence loss for the embedding to conform it to a unit Gaussian. 

Using VAEs for representation learning, followed by a DRL control policy is not new, see for instance \cite{Ha18, Nair18}. The VAE is generally pre-trained, and so the perception is learned independent of control. In \cite{Splitnet19}, the authors used a vanilla Autoencoder (AE) to compress the image, without the KL divergence loss. One difference between the two approaches is the that when VAE is used for perception, the resulting embedding is stochastic, i.e., the same input image fed to the encoder multiple times will result in different embeddings as they are sampled from the Gaussian; whereas in the vanilla AE this is not the case. Other approaches such as using shared latent spaces \cite{liu2017unsupervised, genslam} can also be considered in future studies. 

Another challenge in vision-based robot navigation is on transfer learning to new targets and/or new scenes. To this end, \cite{AI2THOR} trained RGB-based DRL navigation policies for one or more scenes and used this to transfer learn (or fine-tune) to newer scenes. They showed that the DRL learns faster and the overall trajectory length is shorter if more scenes are used in the training. This improves the overall data-efficiency of the DRL training on newer scenes/targets. We will also briefly address transfer learning in the experiments conducted in this paper (section \ref{experimental_results}).

\subsection{Robot navigation with A* and vision}
 
Robot navigation using vision-based sensors is gaining renewed interest in the literature with the advent of state-of-the-art simulators \cite{Habitat19, Gibson, AI2THOR}, robust datasets \cite{Gibson, Matterport}, and efficient deep learning algorithms. A goal-driven DRL framework for visual robotic navigation was provided by \cite{Zhu16}. Robot navigation using a PointGoal, i.e., the position of the goal with respect to the current location of the robot/agent, was used in \cite{Anderson18, Habitat19}; we undertake the PointGoal navigation task in this study, but with a curriculum that gradually increases the difficulty during training. A hierarchical method for navigation combining a sampling based path planning with DRL, called PRM-RL, was proposed in \cite{prmrl}. Their DRL agents were trained for short-range, point-to-point navigation capturing robot dynamics and task constraints without knowledge of the large-scale topology, while Probabilistic Roadmaps (PRMs) as sampling-based planners were used to provide roadmaps which connect robot configurations. Our hybrid path-planning/DRL is similar in spirit to theirs, but we use A* as our path-planner and use a curriculum-based training of the DRL agents.

Specifically, we undertake an investigation of using a small number of waypoints between the start and target locations for the DRL agent to successfully learn to navigate, as well as aiming for longer start-to-target distances as the learning progresses (more details on this in Section IV). Note that we are sequentially increasing the complexity of the navigation problem on the policy learned by the DRL agents, sticking to a pre-determined curriculum. Our approach is not imitation learning as the actions have to be learned by the DRL agents by exploration. Furthermore, despite the use of a two-level hierarchy for the policy, i.e., A* for waypoint determination (in training only) and a DRL policy for the navigation action, we are not undertaking Hierarchical Reinforcement Learning (HRL) for navigation in the spirit of \cite{HAC}. HRL involves learning at multiple levels, whereas in our case the higher-level A* is a graphical search algorithm, and only the lower-level DRL policy involves learning from data.

\section{Problem Setup}
\label{problem_setup}

\textbf{The PointNav task}
We use the Habitat simulator \cite{Habitat19} to train our DRL to learn policies for the point-goal navigation task in the Gibson environment \cite{Gibson}. The robot/agent is equipped with an RGBD camera, a point-goal sensor and a heading sensor. The point-goal sensor is like an indoor GPS: it provides the agent with its current position and the relative position of the target location. The heading sensor provides the current global heading angle of the agent. In the point-goal navigation task, the agent is asked to navigate from the initial starting position to the required end position using only its RGBD, heading and point-goal sensors and without a map. These start and target locations are randomly initialized at the beginning of each episode, for which no straight line path is possible. The agent needs to learn navigation strategies that avoid obstacles and negotiate doorways since the start and target locations can be in different rooms. 

\textbf{Twin-VAE}
We pre-train perception in the environment by using a twin-VAE setup as shown in Figure \ref{fig:vaeRL}. RGBD cameras are initialized randomly in the environment, and at each location, RGB and depth images are collected at angular increments of 10$^o$ for a full 360$^o$ sweep.
These images are used to train the RGB and depth encoder-decoder branches (blue and purple in the figure) with the standard VAE reconstruction and KL divergence losses \cite{kingma2013auto}. Once the VAE is pre-trained, only its encoders are used for training the DRL. RGB and depth images are encoded to their respective embeddings, which are concatenated to provide the final visual embedding from the camera. This embedding is used for training the DRL policy.

\textbf{A* Curriculum Learning}
The task of learning the DRL policy is assisted by incrementally increasing the difficulty of the PointNav task. We do this during training by using A* to determine an optimal path between start and target locations in the bird's eye view (BeV) map of the environment. A new sub-goal, a point to navigate to, that is on this A* path, is provided to the DRL. This sub-goal is close to the starting location to begin with, and then as training progresses, gets farther and farther away from it. We test the following variants of curriculum learning based on discrete and continuous subdivisions of the path:
\begin{enumerate}
\item{\textbf{WP-N}}: In Way-point-N or WP-N, the A* path is divided into N equidistant waypoints (WPs) including the target location. At the beginning of the training episode, the agent is asked to navigate to the first WP. When it reaches within 0.2m of this WP, the goal is revised to the next one and so on till the final target location. We investigate the number of intermediate waypoints required for successful navigation by experimenting with WP-10, WP-8, WP-6. WP-4, WP-3 and WP-2. WP-1 involves no subdivisions of the path and is the same as the original PointNav task.
\item{\textbf{SWP-N}}: Sequential WP-N or SWP-N involves keeping the number of WPs constant for a fixed number (few thousand) episodes. This is the same as WP-N, where the agent is asked to navigate from the 1st to the Nth waypoint within the same episode. However, N decreases episodically. Once the agent has mastered a higher N, requiring a smaller length sub-path traversal, the agent is subjected to a lower N, requiring a larger length sub-path traversal. For instance, the start-to-target path is divided into N waypoints for every 10k episodes, and N is decreased following the set: (10, 8, 6, 4, 3, 2, 1). Note that WP-1 is the same as PointNav. 
\item{\textbf{FWP}}: Farther Waypoint involves only one WP that moves farther and farther away from the start in continuous, linear increments, as training progresses. The training is commenced with the WP at 20\% distance along the A* path from the start. Over the course of training, this WP is moved farther and farther along the path until it is at 100\% of the distance from the start to target after several tens of thousands of episodes, at which point the FWP problem is the same as the PointNav problem.
\end{enumerate}
Note that SWP-N and FWP become PointNav agents by the end of their training. 
However WP-N is an agent that has only mastered a shorter navigation distance compared to the original PointNav agent. We use WP-N to set up the problem and demonstrate the efficacy of navigating smaller paths and using this to boot-strap the training of longer path (SWP-N and FWP) agents. 
Hence, SWP-N and FWP do not require A* at test time, whereas WP-N agents require A* at test time to obtain intermediate goal locations. In this paper, we will focus more on SWP-N and FWP for this reason.

% In this study, we tackle the problem of indoor robot navigation by separating visual perception and control policy. We also slowly increase the difficulty of the navigation task that the control policy has to solve. The robot has access to four sensors: (1) an RGB camera that provides an RGB color image of the immediate field-of-view of the robot; (2) a Depth camera that provides a Depth map of the environment from the robot-centric view; (3) a PointGoal sensor which measures the dynamic geodesic coordinates of the goal with respect to the current location of the robot and is a vector of two real numbers; (4) a heading sensor that measures the heading angle of the robot. The initial position and final goal are randomly generated by the Habitat simulator, which inherently avoids an initial/goal pair for which a straight-line route is possible \cite{Habitat19}; this makes the problem more challenging.     

At any time instant $t$, let the sensory readings be denoted as follows: (1) RGB image, $s^{RGB}_t$; (2) Depth map, $s^{Depth}_t$; (3) pointgoal sensor reading, $PG_t$; and (4) heading angle, $H_t$. The pre-trained twin-VAE is used to encode $s^{RGB}_t$ to $z^{RGB}_t$ and $s^{Depth}_t$ to $z^{Depth}_t$. These are concatenated with the other two sensor readings to obtain a compact representation of the state at time t as: $s_t$ = ($z^{RGB}_t, z^{Depth}_t, PG_t, H_t$). 

% Since the RGB and Depth maps have excess information, an efficient representation learning is warranted to make them compact for faster learning. For visual navigation, many sub-regions of the image are not useful for learning to navigate a robot, and so using a VAE embedding as input to the DRL navigation policy will ensure that only relevant and useful features from the visual sensory measurement are passed on to the policy network for learning to navigate. This speeds up the overall training of the navigation policy. Moreover, the visual sensory inputs are the same for a robot for different tasks in the same environment, and so by separating perception and control, one can train the perception network only once and re-use it multiple times for different navigation tasks; see \cite{WatkinsValls, Splitnet19} for more discussion along these lines.

%Jay: removed above para as embedding implicitly removes unwanted features from the image. This explanation is un-necessary.
%TODO: add re-use of perception network discussion and \cite{WatkinsValls, Splitnet19} earlier, in the introduction.

Once the sensory readings are concatenated into a compact $s_t$, we use Deep Reinforcement Learning (DRL) to learn a policy $\pi_{\theta}$ that outputs action $a_t$ at time t:

\begin{equation}
a_t = \pi_{\theta}(s_t),
\end{equation}
where the actions are one of three: (1) move forward by 0.25 m; (2) turn left by 10$^o$; (3) turn right by 10$^o$. A fouth action called ``Done" is executed whenever the agent is within 0.2 m from the goal position. Note that in \cite{Habitat19}, they trained an agent to also learn this trivial task, but this was not the case in \cite{WatkinsValls, Splitnet19}; we take the latter approach. A schematic of the overall setup is shown in Fig. $\ref{fig:vaeRL}$.

% In addition, we also investigate the use of A* to obtain high-level waypoints following a curriculum, and use them as sub-goals to learn a navigation policy using DRL. Specifically, we use a low-resolution map of the environment to learn an A* path, and use a fixed number of equidistant waypoints from this A* path to revise the goals for the DRL agent, thereby simplifying the learning involved in navigation. Furthermore, we also undertake a principled approach to investigate the effect on learning of the number of waypoints used, as well as to sequentially complicate the agent's learning by (1) choosing fewer number of waypoints with advances in learning; (2) to make the waypoint farther with learning advancement. These ideas are borrowed from curriculum learning \cite{curriculum} where simpler tasks are learned at the beginning, and the level of difficulty is sequentially increased depending on the progress made.

%Jay: TODO: Combine parts from Problem definitions section into this Problem Setup section.

%\begin{figure}[t]
%\begin{center}
%    \includegraphics[width=0.98\linewidth]{images/VAERL.png}
%\end{center}
%   \caption{The encoder part of the pre-trained twin-VAE is used during training the RL policy. This speeds up the training of the RL algorithm.}
%\label{fig:VAERL}
%\end{figure}

% \begin{figure}[t]
% \begin{center}
%     \includegraphics[width=0.98\linewidth]{images/rlfig.png}
% \end{center}
%   \caption{Schematic of the coupled VAE and RL system.}
% \label{fig:RLfig}
% \end{figure}

\section{IMPLEMENTATION DETAILS}
\label{implementation_details}
\subsection{Training the VAE}

The RGB and Depth maps obtained from Habitat are 256$\times$256 and are resized to 192$\times$192 before being fed into the VAE for training. For the Encoder of the VAE, we use 10 layers of 4$\times$4 filters with the number of feature maps per layer varying from 64 to 512. This is followed by fully-connected layers of size $N_z$ each to obtain $\mu_z$ and log $\sigma_z^2$. The reparameterization trick is used to obtain z \cite{kingma2013auto}, the Encoder's embedding vector. For the Decoder, we use a mirror image of the Encoder to scale back to 192$\times$192 size. For the latent code, we will consider a size of $N_z$ = 128 and 256 in this study. All layers use the Relu activation function, except the final output layer of the Decoder which uses a Sigmoid; the intermediate latent layer uses no activation functions to determine $\mu_z$ and log $\sigma_z^2$ for the Gaussian latent code. We use a batch size of 64 and 50,000 iterations to train each of the VAEs, and use the Adam optimizer \cite{adam}.

\subsection{Training the DRL Policy}

The Proximal Policy Optimization (PPO) algorithm \cite{PPO} is used to train the policy network. 
The input to the DRL algorithm comprises of the visual embedding plus the point-goal and heading data as detailed in the previous section (This state at time $t$ is given by $s_t$ = ($z^{RGB}_t, z^{Depth}_t, PG_t, H_t$).
The policy network consists of two fully connected layers with 512 and 256 units and tanh activation function, followed by a LSTM layer \cite{lstm} with 256 units. The output of the LSTM branches out into the policy and value streams, each going through a fully connected layer with 256 units and the tanh activation function. This is followed by a Softmax layer with 3 probabilities corresponding to the actions described earlier 
%Jay: (TODO: Add section cross-ref).   
%Jay TODO: Tweak above para slightly for non-RL audience.

For training the DRL agent, the reward function at time $t$, $r_t$, is same as \cite{Habitat19} and is given by:

\begin{equation}
r_t = \left\{\begin{matrix}
S + d_{t-1} - d_{t} + \lambda \,\,\,\,\, \text{goal reached} \\ 
d_{t-1} - d_{t} + \lambda \,\,\,\,\, \text{otherwise}
\end{matrix}\right.
\end{equation}
where $d_t$ is the distance to the goal from the agent's current location at time $t$, $S$ = 10 is a bonus for reaching the goal, and $\lambda$ = -0.01 to penalize a stationary agent. The Adam optimizer \cite{adam} is used to train the policy network as well. Other PPO-related hyper-parameters are selected as follows: discount factor $\gamma$ = 0.95, PPO clip value $\epsilon$ = 0.1, and Generalized Advantage Estimation \cite{GAE} parameter $\lambda$ = 1.0.

\subsection{Evaluation metrics}

Similar to other Habitat-based navigation work, \cite{Habitat19, Splitnet19, WatkinsValls}, we use the Success-weighted Path Length (SPL) for evaluating the policy learnt at the end of each episode:

\begin{equation}
SPL = S\, \frac{l}{\text{max}(l,p)}    
\end{equation}
where $S$ is a binary indicator of success, $l$ is the shortest path length from start to goal position and $p$ is the path length traveled by the agent in the episode. We also evaluate the Mean Success Rate, which is the mean of $S$ over a fixed number of episodes.

\section{EXPERIMENTAL RESULTS}
\label{experimental_results}

Our primary interest in this study is to undertake indoor robot navigation in the Habitat environment \cite{Habitat19}. The {\it Quantico} indoor scene in the Gibson \cite{Gibson} dataset is used for the experiments, except for the Transfer Learning (TL) experiments, where we TL from {\it Quantico} to the {\it Pleasant} environment. Our experiments relate to the VAE latent encoding parameters, followed by quantitative and qualitative evaluations of the baseline PointNav agent compared to the agents (WP-N, SWP and FWP) trained by A* curriculum learning.
%Jay: We need to consolidate sections B and C.
%Jay TODO: Add timing information and show the ramifications of A* curriculum learning for practical deployment of robots. Something like... training a PointNav agent on a 2080Ti GPU is 2k episodes per day, and requires 5 days for 10k episodes. A speedup of a factor of 4 (20k vs 80k episodes for FWP compared to Point-Nav) means the agent takes 5 days to learn instead of 20 days on a high-end desktop GPU. This has significant ramifications for training DRL policies for mobile robotics.

\subsection{Choice of latent encoding}

\begin{figure}[h]
\begin{center}
    \includegraphics[width=0.85\linewidth]{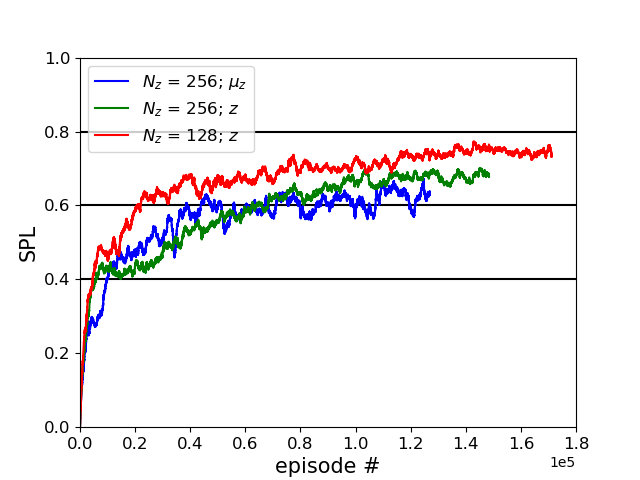}
\end{center}
   \caption{Effect of the latent encoding on the training of DRL agents.}
\label{fig:enc}
% \vspace*{-0.80cm} 
\end{figure}

As aforementioned, for representation learning, we use separate VAEs to encode the RGB and Depth images, concatenating the two encodings for subsequent use to train the DRL agents. For this we can use either the latent code $z$ sampled from the Gaussian $z \sim p(z)$, or we can directly use the mean of the Gaussian $\mu_z$. For instance, \cite{Ha18} uses the VAE's $z$ for the RL, whereas \cite{Nair18} uses $\mu_z$. Likewise, the dimension of the latent code can also have an effect on the training of the DRL agents. To better understand the choice of the VAE encoding that would result in a robust encoding of the visual inputs for efficient training of the DRL agents, we consider three cases: (1) $N_z$ = 256; $\mu_z$, (2) $N_z$ = 256; $z$, and (3) $N_z$ = 128; $z$. Separate DRL agents are trained on the PointNav problem using these three different choices and the exponentially-averaged (with degree of weighting $\alpha$ = 0.001) SPL during the training of the agents are presented in Fig. $\ref{fig:enc}$. As evident, $N_z$ = 128 performs better than 256, which we believe is due to over-fitting when the latent code dimension is large. Likewise, the training is better when $z$ is used in lieu of $\mu_z$, as the stochasticity involved in $z$ acts as a regularizer and can be a good source of exploration noise. For the rest of this paper, we will use $N_z$ = 128 and $z \sim p(z)$ as the encoding for training the DRL agents.  
%Jay: Can we say that using the mean of the Gaussian, mu(Z) is equivalent to using an Auto-encoder without the generative, i.e., VAE part?
%This would show through experiments that the VAE is a better choice than the AE.
%KB: No, it isn't, because we are still using the KL divergence loss during training.
% Using mu(z) during testing means that we are taking away the stochasticity of the encoder.

\subsection{Navigation results}

We will now summarize results from the different cases of interest as identified earlier. The best performing PointNav agent is used as the baseline (this is the agent corresponding to the red curve in Fig. $\ref{fig:enc}$). Note that we will only compare this PointNav baseline performance with SWP and FWP agents as these two agents are essentially PointNav at the end of the training. Performance of the WP-N agents are presented only for demonstration that shorter paths lead to improvement in learning; note that WP-N agents are not desired for deployment as they still require the help of A* at test time. Thus, the test time performance of WP-N agents are not presented or compared with other agents as we are only interested in agents that use A* in training but not at test time.

\subsubsection{PointNav}

For the PointNav problem, the best performing agent is the $N_z$ = 128 and $z$ sampled from the Gaussian $z \sim p(z)$, i.e., the agent corresponding to the red curve in Fig. $\ref{fig:enc}$. We test the agent's performance over 500 random test episodes. The only difference between training and test mode is that in training the action is sampled from the policy's softmax output, but in test mode, the greedy action is chosen. The best performing PointNav agent (red curve in Fig. $\ref{fig:enc}$) has a test time performance of mean SPL = 0.73 and mean success rate = 0.852. We will use these values as the baseline for comparing the other agents with. 

\subsubsection{WP-N}

\begin{figure}[h]
\begin{center}
\centering
    \includegraphics[width=0.85\linewidth]{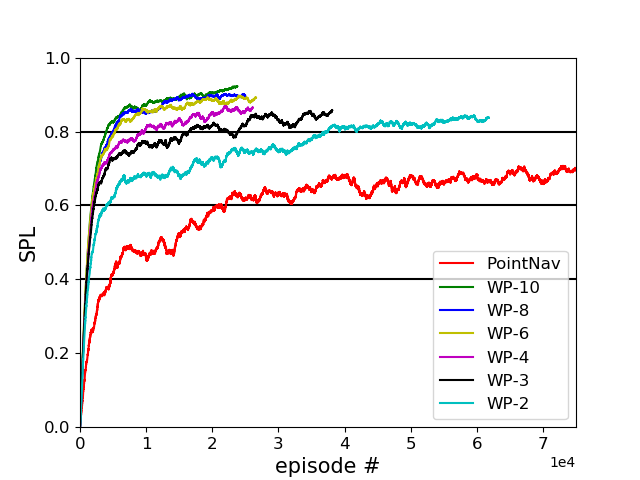}
   \caption{Training performance for the WP-N agents. The PointNav agent (red curve) was run till $\sim$ 170k episodes (see the red curve from Fig. $\ref{fig:enc}$) but is presented here only till 75k episodes for better comparison with the other agents presented in the figure.}
\label{fig:wpn}
\end{center}
\end{figure}

We present the training curves for the different WP-N cases considered, along with the best performing PointNav agent in Fig. $\ref{fig:wpn}$. For WP-N, the SPL is computed as a piecewise average over the individual sub-goals. The availability of the intermediate waypoints has significantly improved the performance for the WP-N cases vis-\'a-vis the PointNav case (refer to Fig. $\ref{fig:wpn}$), as the overall problem of navigation from start to goal has been simplified. With fewer number of intermediate waypoints, the agents take longer to achieve an average SPL of over 0.8, which is as expected. 

The averages of the SPL and the success rate of the agents for 500 test episodes are summarized in Table $\ref{tab:wpn}$. The success rate is over 0.98 and the SPL over 0.85 for all the WP-N agents but not the PointNav agent. This investigation proves that only a small number of intermediate waypoints suffices, and a DRL policy can make efficient use of it to learn to navigate much faster than the PointNav cases. This has important implications for robot navigation applications, as a small handful number of intermediate goal points to visit between the start and target can be generated even with a low resolution map, and one can then combine it with a DRL-based navigation policy and still obtain high accuracy. From Fig. $\ref{fig:wpn}$, we observe that even 1-2 intermediate waypoints (WP-2 and WP-3) between the start and target positions suffices. However, as aforementioned, these WP-N agents are not preferred for deployment/test time as they still require the use of A*. We will now consider the hybrid training practice of using WP-N agents in training and following a curriculum (i.e., SWP and FWP agents) to gradually transition them to the PointNav agents so that they can be deployed without any further use of A* at test time.

\begin{center}
\begin{table}[]
\centering
\begin{tabular}{|c|c|c|}
\hline
Case name & SPL  & Success rate \\
\hline
PointNav  & 0.73 & 0.852        \\
WP-10     & 0.92 & 0.998        \\
WP-8      & 0.9  & 0.992        \\
WP-6      & 0.9  & 0.994        \\
WP-4      & 0.88 & 0.984        \\
WP-3      & 0.88 & 0.983        \\
WP-2      & 0.85 & 0.988       \\
\hline
\end{tabular}
\caption{SPL and success rate for WP-N agents}
\label{tab:wpn}
% \vspace*{-0.5cm}
\end{table}
\end{center}

\subsubsection{SWP-N and FWP}

Having established that a few intermediate waypoints between the start and target locations suffice, we will now investigate the SWP-N and FWP problems, both of which involve curriculum-based training. Note that WP-N agents require waypoints at both training test time, whereas the SWP-N and FWP agents require intermediate waypoints only during training, and are thus preferred. For SWP-10, we follow the curriculum: WP-10($<$10k), WP-8(10-20k), WP-6(20-30k), WP-4(30-40k), WP-3(40-60k), WP-2(60-80k) and WP-1($>$80k), where the number in parenthesis denotes the episode number range. (Note: WP-1 is identical to PointNav). For the FWP agent, we start the training with the revised goal set as the 20-th percentile of the start-to-goal A* path, and gradually increase it linearly over the course of 64k episodes to the 100-th percentile of the A* path. The FWP problem also reduces to the PointNav problem at the end of the training as the waypoint the agent is shooting for coincides with the target location (for episode number $>$ 64k). 

\begin{figure}[h]
\begin{center}
    \includegraphics[width=0.85\linewidth]{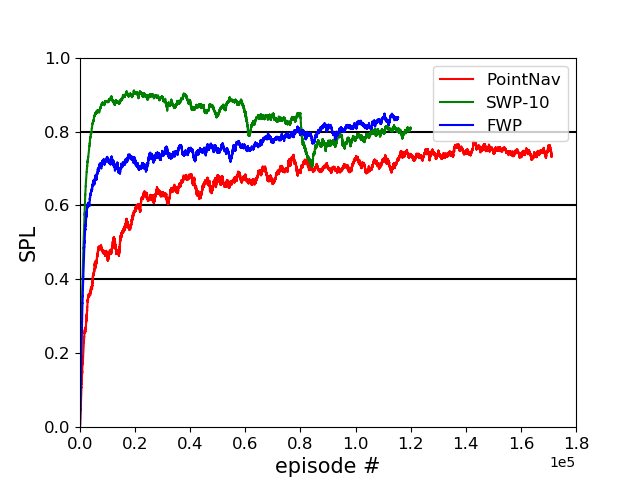}
\end{center}
   \caption{Training performance for the SWP-10 and FWP agents.}
\label{fig:fwp}
% \vspace*{-0.75cm} 
\end{figure}

The training curves are presented in Fig. $\ref{fig:fwp}$ for the SWP-10, FWP and the baseline PointNav agents. We observe that SWP-10 agent achieves high SPL values in the early stages of the curriculum, but drops in performance as the curriculum is more difficult at the later stages of the training. On the other hand, the FWP agent has learned to achieve SPL values of $\sim$ 0.8 and maintains the same level of performance. Both the SWP-10 and FWP agents (both of which are essentially PointNav at the end of the training) maintain superior performance over the PointNav agent. Furthermore, both SWP-10 and FWP agents also learn much faster than the PointNav agent. 

The averages of SPL and success rates over 500 test time episodes for the SWP-10 and FWP agents are summarized in Table $\ref{tab:fwp}$. Note that at test time, the greedy action from the policy probabilities is used, instead of sampling. Both SWP-10 and FWP agents maintain a success rate of over 0.9 and the SPL is also higher for these agents compared with the PointNav agent. The SWP-10 agent performs the best among all the agents and achieves an average SPL of 0.91. The SWP and FWP agents also learn much faster than the PointNav agent (see Fig. $\ref{fig:fwp}$). Thus, learning with a curriculum assists in learning a better policy and also faster (measured in terms of number of training episodes).

\begin{center}
\begin{table}[]
\centering
\begin{tabular}{|c|c|c|}
\hline
Case name & SPL  & Success rate \\
\hline
PointNav  & 0.73 & 0.852        \\
SWP-10     & 0.91 & 0.96        \\
FWP      & 0.77  & 0.9        \\
\hline
\end{tabular}
\caption{Test time SPL and success rate for SWP-10 and FWP agents}
\label{tab:fwp}
% \vspace*{-0.6cm} 
\end{table}
\end{center}

\subsection{Comparison of different agents}

\begin{figure*}[h]
\centering
\begin{tabular}{c c c c}
\subfloat[] { \includegraphics[scale=0.65]{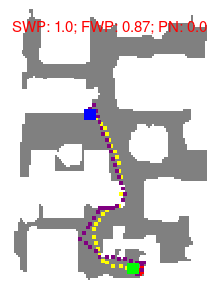} } & \subfloat[] { \includegraphics[scale=0.65]{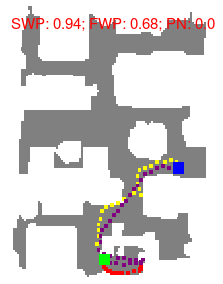} } & \subfloat[] { \includegraphics[scale=0.65]{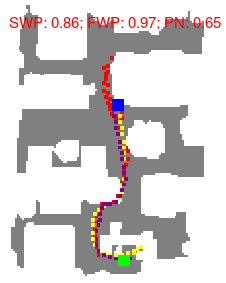} } & \subfloat[] { \includegraphics[scale=0.65]{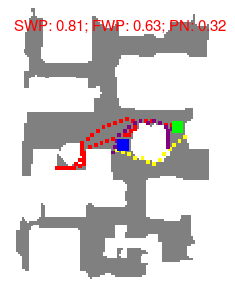} } \\ 
\subfloat[] { \includegraphics[scale=0.65]{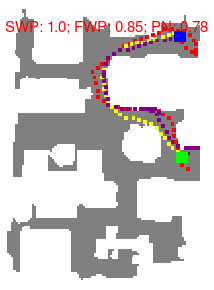} } & \subfloat[] { \includegraphics[scale=0.65]{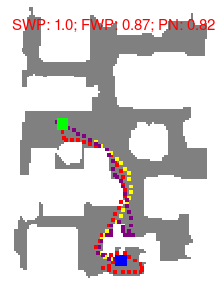} } & \subfloat[] { \includegraphics[scale=0.65]{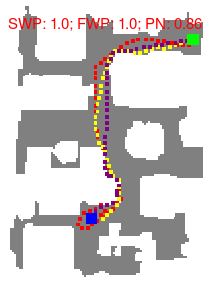} } & \subfloat[] { \includegraphics[scale=0.65]{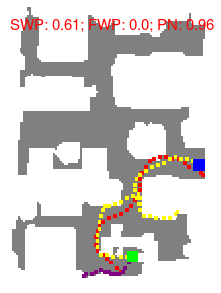} } 
\end{tabular}
\caption{Test time paths traced by the PointNav, SWP-10 and FWP agents for different episodes. The start and target positions are represented by the green and blue squares respectively. The paths traced are shown in red for PointNav, yellow for SWP and purple for the FWP agents. The SPL values for the three agents for the respective episode are also shown in each sub-figure. We recommend that this figure be viewed in digital format and zoomed in for better clarity.}
\label{fig:compare}
% \vspace*{-0.4cm} 
\end{figure*}

We now compare the test time performance of the PointNav, SWP-10 and FWP agents for a few test episodes. The paths traced by the agents, including the start and target positions, for 8 episodes at test time are shown in Fig. $\ref{fig:compare}$. The observations reported here are based on performance over several more test episodes, but we will discuss only these 8 chosen test episodes for brevity; we will use the term ``episode-a" to refer to Fig. $\ref{fig:compare}$ (a) and so on for the other episodes as well. In episodes-a and b, the PointNav (shown in red) fails to reach the target position. We noticed many episodes where the PointNav agent failed closer to the start position than otherwise; this agent has starting trouble in such episodes. On the other hand, agents SWP and FWP successfully navigated to the goal for episodes-a and b, with the SWP agent being smoother for episode-a and FWP for episode-b. 

In many episodes, we also observed the PointNav agent to reach close to the target, only to overshoot it and get confused, but the agent was still able to reach the target, albeit with an unnecessarily longer path. This is reflective in episodes-c, d, e, and f (notice in the vicinity of the target, shown by the blue square). The SWP and FWP agents perform better in these episodes and successfully reach the target in shorter paths than the PointNav agent. 

The FWP agent is coasting more than the SWP, as evident from episodes-a, e, and f, where we observe the FWP agent to stay closer to walls and make near 90$^o$ turns. In these episodes, the SWP agent has relatively smoother paths, which we desire for real word deployment. However, in episodes-b and g, the reverse is observed--the SWP agent (shown in yellow) is coasting more than the FWP agent. This is an interesting behavior in that one agent is not ``better" than the other agent in all test episodes, and the location of the start and target positions has an influence on the agents' performance. This behavior is observed in other episodes as well (not presented in the Figure), and the choice of the curriculum used in the training has a significant influence in the final test time performance. This has implications to real world robots that are trained using a DRL-based navigation policy following a curriculum. 

In episode-h, the PointNav agent actually performs better than the other two agents, which is a very rare outcome. For this episode, the FWP agent (in purple) actually fails to reach the target and is stuck in a corner. Another interesting observation is that episodes b and h have start locations somewhat close to each other, as well as target locations, but the FWP agent (shown in purple) is able to successfully navigate episode-b but not episode-h. This, however, is quite rare an occurrence from our observation of the other test episodes as well (not presented here). Thus, agents can fail for even a slight increase in the level of difficulty at times.

 \subsection{Transfer Learning (TL) Experiments}

One of the test of robustness of a learned policy on one task is its ability to generalize to another, but related, task. To this end, we undertake TL of VAE representations and DRL policies from {\it Quantico} to the {\it Pleasant} scene in the Gibson dataset \cite{Gibson}. Both scenes correspond to indoor homes with living rooms, bedrooms, etc., and have similar, albeit not same, furniture/appliances. Similar tests were also undertaken by \cite{AI2THOR} for evaluating TL for robot navigation. The VAE is fine-tuned for {\it Pleasant} using the pre-trained VAE weights from {\it Quantico}. The SWP-10 trained for {\it Quantico} is used as starting policy network weights and fine-tuned on {\it Pleasant} following a PointNav approach (i.e., without any A* curriculum), and this agent after transfer learning (TL) on the new scene is referred to as TL-SWP. 
After re-training for $\sim$ 60k episodes on the new scene, on 500 test episodes, TL-SWP has a mean SPL of 0.87 and a mean success rate of 0.95. Thus, with fewer episodes of re-training on the new scene, TL achieves reasonable performance. In future work, we expect to conduct a more systematic TL investigation on Habitat scenes similar to \cite{AI2THOR}, where the number of scenes used in the first training was shown to have an impact on the transfer learning performance.

\section{CONCLUSIONS}
\label{conclusions}
We look at the task of point-to-point navigation (PointNav) in the Habitat environment \cite{Habitat19}. An agent gets photo-realistic RGBD images from this environment and learns an optimal navigation policy using Deep Reinforcement Learning (DRL). We introduce two ways of improving, in terms of speed and final metrics, the DRL training process: (1) We separate the perception and control tasks by pre-training a VAE to learn an embedding for the RGBD data. (2) We use A*, a traditional robotic path planning algorithm like training wheels on a bike that eventually come off, and train the DRL algorithm in an incremental curriculum. This involves increasing the path length required to successfully complete the PointNav task. We experiment with two curricula: one that increases the path length in a continuous fashion, and another that does this discretely. At test time, we do not need the help of A* any more. 

% A* provides a series of waypoints between randomized start and target locations, and use a varying number of these to help the DRL algorithm, in addition to the inherent DRL reward function. Our DRL is trained in a curriculum fashion, presenting the agent with a series of path planning tasks that incrementally increase in difficulty levels in terms of path distance. 

Our agents learn faster compared to the PointNav baseline as well as achieve higher final SPL scores and success rates. The perception embedding is trained once for an environment, and can then be used for a variety of agent policies. The training of a number of sub-policies allows these policies to be used independently for shorter-distance navigation, or to be used to build up more sophisticated long range navigation policies. The ability to scan a real-world environment and set up its photo-realistic virtual twin for training direct perception-to-navigation policies is expected to accelerate the deployment of mobile robots around the homes and factories of our future. Our work is a step improvement in this direction.

In future work, we plan on using conventional local path planning approaches to augment the reward function for DRL and try to imbibe the network with a neural map \cite{Chaplot20}.

% We present two training curricula: (1) The Sequential Waypoint (SWP) method, which uses a discrete subdivision of the path output by the A* algorithm; and (2) The Farther Waypoint (FWP) method that is its continuous counterpart. Both SWP and FWP agents start with small path-lengths along the A* route to be navigated, but by the end of the training, are navigating the full paths between the start and target locations without any further assistance from A*. Our simulations show that the SWP and FWP agents perform better than the baseline PointNav agent, with the SWP agent having overall best performance. These agents trained in a curriculum fashion also learn much faster compared to the PointNav baseline.

%Jay: trying to shorten conclusion section, removed above para.

% The ability to scan real-world environments, set up their virtual twins in simulation, train direct perception to navigation policies and subsequently deploy them in the real world, is expected to accelerate the adoption of mobile robots. Our work is a step improvement in this direction. 

% In future work, we plan to investigate other curricula for improving the training process. We also plan on using conventional local path planning approaches to augment the reward function for DRL and try to imbibe the network with a neural map \cite{Chaplot20}.
% Jay: removed future work to save space.

\FloatBarrier

% \addtolength{\textheight}{-4cm}  

\bibliographystyle{IEEEtran}

\bibliography{main}

\end{document}